\documentclass[runningheads]{llncs}
\usepackage{graphicx}

\usepackage{tikz}
\usepackage{comment}
\usepackage{amsmath,amssymb,amsfonts} 
\usepackage{color}
\usepackage{xcolor}

\usepackage[accsupp]{axessibility}  

\usepackage{amsmath}
\newcommand{\eg}{\textit{e.g.,\ }}

\begin{document}
\pagestyle{headings}
\mainmatter

\title{\raisebox{-0.16cm}{\includegraphics[scale=0.14]{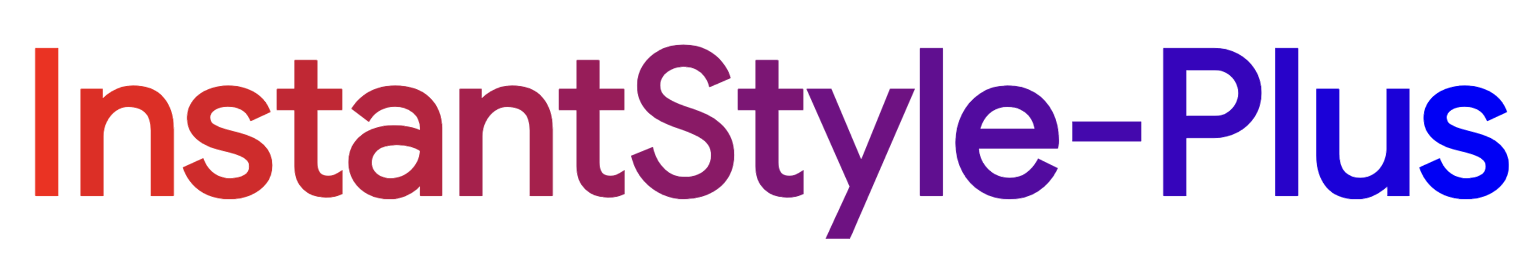}}: Style Transfer with Content-Preserving in Text-to-Image Generation} 

\titlerunning{InstantStyle-Plus: Style Transfer with Content-Preserving}

\author{Haofan Wang \and
Peng Xing \and
Renyuan Huang \and
Hao Ai \and \\
Qixun Wang \and
Xu Bai
}

\authorrunning{Wang et al.}
%


\institute{
InstantX Team \\
\email{\{haofanwang.ai@gmail.com\}\\
\textcolor{magenta}{\url{https://instantstyle-plus.github.io}}
}
}
\maketitle

\begin{abstract}

Style transfer is an inventive process designed to create an image that maintains the essence of the original while embracing the visual style of another. Although diffusion models have demonstrated impressive generative power in personalized subject-driven or style-driven applications, existing state-of-the-art methods still encounter difficulties in achieving a seamless balance between content preservation and style enhancement. For example, amplifying the style's influence can often undermine the structural integrity of the content. To address these challenges, we deconstruct the style transfer task into three core elements: 1) Style, focusing on the image's aesthetic characteristics; 2) Spatial Structure, concerning the geometric arrangement and composition of visual elements; and 3) Semantic Content, which captures the conceptual meaning of the image. Guided by these principles, we introduce InstantStyle-Plus, an approach that prioritizes the integrity of the original content while seamlessly integrating the target style. Specifically, our method accomplishes style injection through an efficient, lightweight process, utilizing the cutting-edge InstantStyle framework. To reinforce the content preservation, we initiate the process with an inverted content latent noise and a versatile plug-and-play tile ControlNet for preserving the original image’s intrinsic layout. We also incorporate a global semantic adapter to enhance the semantic content's fidelity. To safeguard against the dilution of style information, a style extractor is employed as discriminator for providing supplementary style guidance. Codes will be available at \textcolor{magenta}{https://github.com/instantX-research/InstantStyle-Plus}.

\keywords{Style Preservation, Content Preservation, Image Synthesis}
\end{abstract}

\section{Introduction}

Recently, diffusion models~\cite{podell2023sdxl,rombach2022high,song2020denoising} have exhibited exceptional capability in the field of image generation, particularly excelling in personalized image generation~\cite{ruiz2023dreambooth,wang2024instantid,ye2023ip} to ensure a consistent style or character. Among these advancements, style transfer has been particularly noteworthy. By incorporating a reference style image, these models adeptly embody the desired style into a series of generated images.

Previous studies~\cite{hu2021lora,ruiz2023dreambooth} have frequently employed fine-tuning of diffusion models on batches of images characterized by a consistent style. However, these approaches require significant computational resources for training or fine-tuning large-scale foundational models. Consequently, they become costly when adapting to new, unseen styles. In response, recent research~\cite{frenkel2024implicit,hertz2023style,jeong2024visual,sohn2023styledrop,wang2024instantstyle,wang2023styleadapter} has pivoted towards the development of tuning-free methods for stylized image generation. The prevalent approach in optimization-free style transfer typically involves manipulation of self-attention or cross-attention layers. For instance, StyleAlign~\cite{hertz2023style} and Visual Style Prompting~\cite{jeong2024visual} leverage inversion techniques and a pre-trained base model to extract intermediate features (keys and values) from the reference style image, subsequently replacing the original features. These methods encounter difficulties in both the extraction of subtle stylistic elements from the reference image and the precise application of these styles onto a target content image through inversion operations. To address these challenges, InstantStyle~\cite{wang2024instantstyle} integrates features from the reference style image into style specific layers, enhancing the style transfer process. However, there is a notable scarcity of works that prioritize content-preserved stylization. More recently, in pursuit of content preserving, several studies~\cite{chung2024style,lin2024ctrl,rout2024rb,xu2024freetuner} have emerged, focusing on enhancing the ability to preserve the integrity of content in stylization. For example, StyleID~\cite{chung2024style} applys initial latent AdaIN and query preservation techinques to mitigate the issue of disruption of original content. FreeTuner~\cite{xu2024freetuner} and Ctrl-X~\cite{lin2024ctrl} operate on self-attention or cross-attention mechanisms, but uniquely, they utilize intermediate features derived from the content image rather than the style image, focusing on the accurate representation of content. Despite of these efforts, existing methods tend to prioritize compositional personalization over in-place content integrity, or they fall short in effectively capturing style, structure, and semantics.

To overcome the challenges in content-preserved style personalization, we introduce InstantStyle-Plus, an enhancement that refines the stylization process while meticulously safeguarding content integrity. Building upon the recent state-of-the-art method, InstantStyle~\cite{wang2024instantstyle}, InstantStyle-Plus integrates a suite of novel techniques specifically tailored to bolster its ability to preserve content. We have meticulously decomposed the task into three distinct yet interrelated subtasks to address the complexity of the challenge: style injection, spatial structure preservation, and semantic content preservation. Correspondingly, we utilize the initial content latent and Tile ControlNet~\cite{zhang2023adding} to underscore their effectiveness for content preservation, We also introduce a global image adapter~\cite{ye2023ip} for semantic retaining. Finally, we adopt the CSD~\cite{somepalli2024measuring} model as our style discriminator, offering additional style guidance to balance content and style.

In summary, we present a pre-experimental report, our primary focus lies in evaluating the utility of established methodologies for content preservation in style transfer tasks, rather than creating a novel framework. We find Tile ControlNet~\cite{zhang2023adding} to be highly effective in maintaining spatial composition, and inverting content noise~\cite{garibi2024renoise} can further enhance the preservation of subtle content details, suggesting that caching intermediate features from content images may be superfluous. We also discuss the often-overlooked semantic component in content preservation, highlighting its utility for maintaining semantic integrity beyond the scope of text prompts. Furthermore, we explore the impact of additional style guidance, contingent upon the effectiveness of a good style discriminator.
\section{Related Work}

\subsection{Stylized Image Generation}

Stylized image generation~\cite{chung2024style,frenkel2024implicit,hertz2023style,jeong2024visual,qi2024deadiff,rout2024rb,sohn2023styledrop,wang2024instantstyle}, commonly referred to as image style transfer, involves the process of transferring the stylistic or aesthetic attributes from a reference image to a target image. A multitude of methods have been developed to ensure style consistency across a series of images generated by diffusion models. In the realm of inversion-based approaches, StyleAlign~\cite{hertz2023style} employs shared self-attention mechanisms with the reference image to align styles effectively. Visual Style Prompting~\cite{jeong2024visual}, on the other hand, retains the original image's query features while integrating the key and value features from the reference image in the later stages of self-attention layers. In contrast, IP-Adapter~\cite{ye2023ip} and Style-Adapter~\cite{wang2023styleadapter} introduce a distinct cross-attention mechanism that decouples the attention layers for text and image features, allowing for a coarse control over the style transfer process. DEADiff~\cite{qi2024deadiff} stands out by extracting disentangled representations of content and style using a paired dataset, facilitated by the Q-Former technique. InstantStyle~\cite{wang2024instantstyle}, a recent innovation, employs block-specific injection techniques to implicitly achieve a decoupling of content and style, offering a nuanced approach to style transfer. However, these studies primarily concentrate on the purity of the style transfer, often overlooking the preservation of the content image's structural integrity.

\subsection{Content-Preserved Style Transfer}

Additionally, another branch of research~\cite{chung2024style,lin2024ctrl,rout2024rb,xu2024freetuner} within the field focuses on style transfer, prioritizing the preservation of the original content's integrity. In essence, these studies concentrate on in-place stylization. StyleID~\cite{chung2024style} operates by manipulating self-attention layers, proposing innovative techniques such as query preservation and initial latent AdaIN to maintain the integrity of the content. FreeTuner~\cite{xu2024freetuner} leverages intermediate diffusion model features for subject concept representation, integrating style guidance to ensure that synthesized images align with both the subject's structural integrity and the style's aesthetic attributes. RB-Modulation advances the field by enabling seamless content and style composition, breaking away from the reliance on external adapters or ControlNets, which is a significant departure in style transfer methodology. Ctrl-X~\cite{lin2024ctrl}, introduces a feed-forward structure control mechanism for precise structure alignment with a reference image, coupled with semantic-aware appearance transfer that facilitates the adoption of user-input appearances, marking a new frontier in customizable style transfer. However, these methods either emphasize compositional personalization rather than in-place content integrity, or are less effective in terms of style, structure, and semantics.

\section{Methods}

\begin{figure}[!h]
  \centering
  \includegraphics[width=\textwidth]{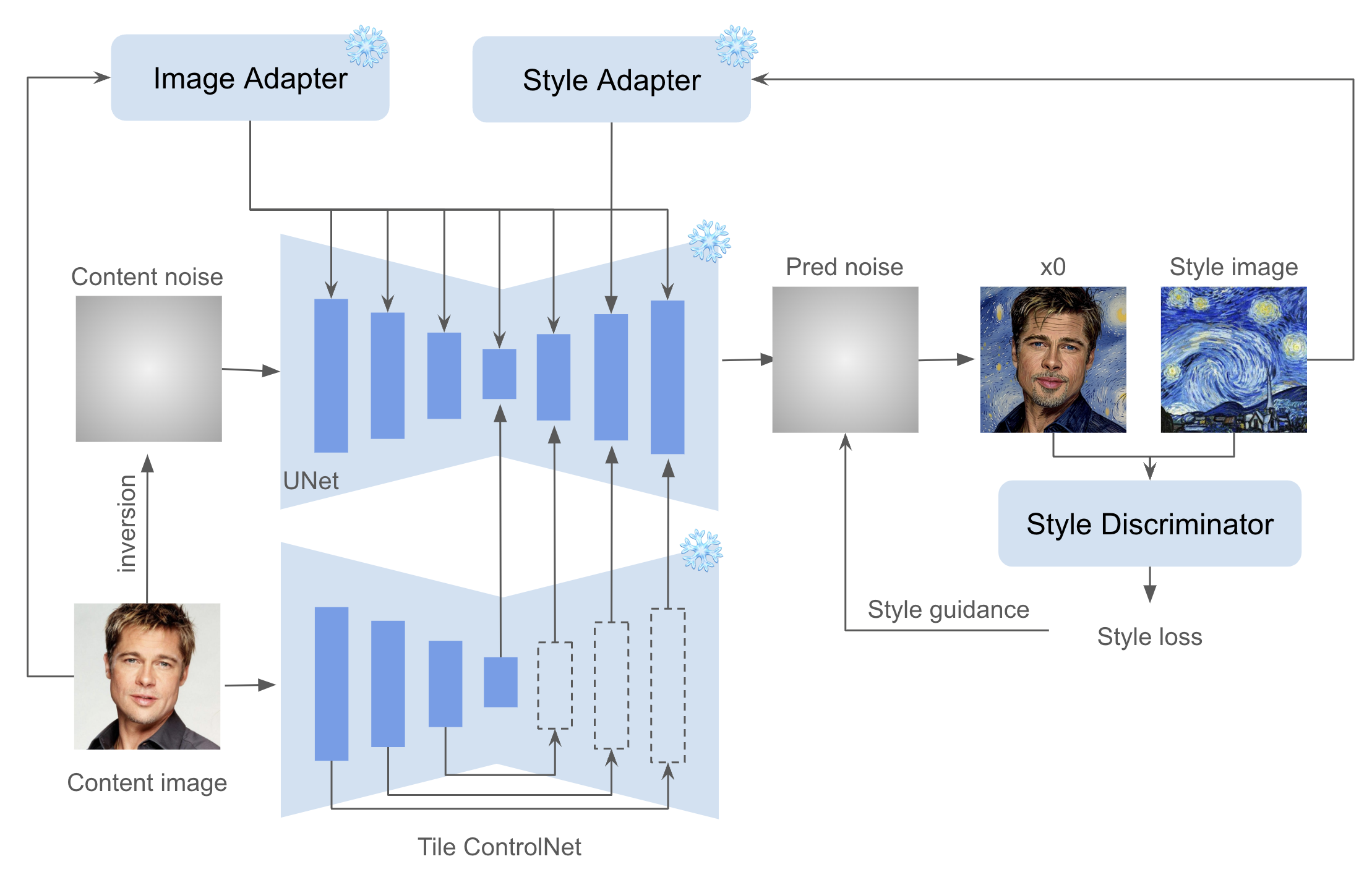}
  \caption{\textbf{Pipeline of our InstantStyle-Plus.} For stylistic infusion, we adhere to InstantStyle's approach by injecting style features exclusively into style-specific blocks. To preserve content, we initialize with inverted content noise and employ a pre-trained Tile ControlNet to maintain spatial composition. For semantic integrity, an image adapter is integrated for the content image. To harmonize content and style, we introduce a style discriminator, utilizing style loss to refine the predicted noise throughout the denoising process. Our approach is optimization-free.}
  \label{overall}
\end{figure}

In this study, rather than enhancing conventional personalized or stylized text-to-image synthesis, we concentrate on a more pragmatic application: style transfer that maintains the integrity of the original content. We decompose this task into three subtasks: style injection, spatial structure preservation, and semantic content preservation. We commence by presenting preliminary experiments and observations, which serve as the motivation for our approach. Subsequently, we delineate our tailored solutions to each constituent subtask in sequence. The overall pipeline of our optimization-free method is shown in Figure \ref{overall}.

\subsection{Preliminaries}

Powered by the generative ability of diffusion models, various personalized or stylized image generation methods (optimization-based or tuning-free) have been proposed to achieve style-consistent text-to-image generation, given a style reference. However, the domain of image-to-image stylization has received scant attention in the literature. In real-world scenarios, readily available stylization methods can seamlessly facilitate image-to-image generation by incorporating additional spatial constraints, such as a plug-and-play ControlNet. In our experiments, we observe that the direct integration of such a module into the generation process can diminish the style strength, resulting in a pronounced trade-off between spatial structure and stylistic effect.

An additional observation is that inversion, a prevalent technique in image editing, has been explored in stylization by prior research. While inversion provides a beneficial starting point for noise, as highlighted in InstantStyle, the resulting inverted style noise tends to omit subtle style nuances, including textural details. Notably, this limitation does not impede our content-preserving task, given that the current inversion techniques sufficiently maintain the overall spatial structure of the original content image. However, inversion alone will still lead to some semantic drift, such as changes in character gender.

In short, we have discovered that achieving a harmonious balance between stylistic effect and content preservation is challenging in current stylization methods. While inversion offers valuable insights, it alone is insufficient to yield satisfactory results. These motivate us to introduce our approach.

\subsection{Style Injection}

We commence our approach with style injection. Similar to numerous style-consistent image generation methods that ensure style alignment across a sequence of generated images, we extend the capabilities of a recent state-of-the-art framework, InstantStyle~\cite{wang2024instantstyle}. This framework leverages cross-attention mechanisms and employs two pivotal strategies to accomplish the disentanglement of content and style. For the majority of commonly known style attributes, we find that an efficient and lightweight image adapter is typically sufficient to address these in text-to-image generation without compromising text controllability.

To this end, we adhere to the implementation principles of InstantStyle. The pivotal strategy involves injecting reference style features selectively into style-specific blocks, employing an implicit decoupling technique. This approach effectively prevents content leakage and eliminates the need for cumbersome weight tuning processes on stylized datasets. Specifically, we employ a decoupled cross-attention strategy as originally proposed in IP-Adapter\cite{ye2023ip}, embedding image features through additional cross-attention layers (a set of Key and Value for image) while leaving other parameters within text-to-image backbone unchanged. The decoupled cross-attention can be illustrated as.

\begin{align}
    Z_{new} = Attention(Q,K^{t},V^{t}) + \lambda \cdot Attention(Q,K^{i},V^{i}),
\end{align}

where $Q$, $K^{t}$, $V^{t}$ are the query, key, and values matrices of the attention operation for text, $K^{i}$ and $V^{i}$ are for image. Given the query features $Z$ and the image features $c_{i}$, $Q=ZW_{q}$
$K^{i}=c_{i}W_{k}^{i}$, $V^{i}=c_{i}W_{v}^{i}$. Note that only $W_{k}^{i}$ and $W_{k}^{i}$ are trainable weights and added into style-specific layers. In addition, unlike the original InstantStyle, we do not explicitly subtract the content features from the style reference in advance for the sake of convenience.

\subsection{Spatial Structure Preserving}

\subsubsection{Initial Content Latent.} Similar to text-to-image, latent initialization is a prevalent technique in image-to-image generation, but in addition to a prompt, an initial image is used as a starting point for the diffusion process. The initial image is encoded to latent space and noise is added to it. The presence of additional noise frequently leads to substantial content drift during the generation process, deviating from our original intent. While in the realm of image editing tasks, rather than relying on auto-encoders for encoding, it is often more expedient to employ image inversion techniques. These techniques reverse the sampling process, yielding a latent noise representation directly from a real image, thereby eliminating the need for additional noise injection. InstantStyle~\cite{wang2024instantstyle} argues that image inversion tends to omit subtle style nuances from image, but this limitation does not impede our content-preserving task. StyleID~\cite{chung2024style} proposes to apply AdaIN on initial
latent of style image and content image to preserve the color tone and  structural information of the content image.

In our approach, style information is injected through cross-attention mechanisms, rendering the use of the initial latent from the style image unnecessary. Specifically, we adopt ReNoise\cite{garibi2024renoise}, a recent inversion approach that performs better than original DDIM inversion~\cite{song2020denoising}, to achieve faithful real content image inversion. The inverted noise is used as initial content latent for denoising. Nonetheless, our analysis indicates that the AdaIN operation, as implemented in StyleID~\cite{chung2024style}, exhibits a degree of triviality in its contribution to the content preserving. Despite its advantages, we have observed that with some real-world images, particularly those human-centered, even sophisticated inversion techniques can result in subtle shifts in both spatial and semantic content, \eg identity and background.

\subsubsection{Tile ControlNet for Content Preserving.} ControlNet~\cite{zhang2023adding} has emerged as one of the most popular techniques~\cite{zhang2023adding,mou2024t2i} for spatial conditionings, \eg canny edge, depth map, human pose and etc. Among these spatial conditionings, the role of tiling is often overlooked. In this work, we demonstrate that tile-based ControlNet applications extend beyond mere upscaling. Unlike other pre-processed conditionings, tile ControlNet works very similar to the structure of ReferenceNet~\cite{hu2024animate}, where the original image without any pre-processing is used as input to genuinely preserve intricate content. In contrast, other conditions tend to omit subtle details from the content image. In our approach, the tile-based ControlNet is implemented on the content image, effectively maintaining the spatial structure in conjunction with the initial content latent. Multi-ControlNets could also potentially offer more precise control, but this aspect is beyond the scope of our current discussion.

\subsection{Semantic Content Preserving}

We have delineated content preservation into spatial and semantic components. Besides of spatial composition, it is also important to preserve semantic content, \eg identity, gender, age and facial expression. A straightforward solution is to provide a detailed textual description of content image, as semantic component is usually easy to describe. In practice, we also employ a global image adapter~\cite{ye2023ip}, which follows the same underlying mechanism as in the style injection process. However, unlike the style-specific block focus in style injection, the global features operate on all blocks. Concurrently, we acknowledge that specialized adapters, such as InstantID~\cite{ye2023ip,wang2024instantid}, are particularly beneficial for enhancing identity preservation. However, for more general applications, a global and coarse adapter suffices for maintaining semantic integrity in addition to text prompt.

\subsection{Supplementary Style Guidance}

In the preceding sections, we emphasized the notable trade-off between maintaining spatial structure and enhancing stylistic effects. Consequently, to bolster the style effect, FreeTuner~\cite{xu2024freetuner} utilizes a pre-trained VGG-19 as an external supervisor to penalize the difference at feature map level between the predicted image and style reference. In this work, we also introduce an additional style guidance in the denoising process as a compensation. In contrast, capitalizing on the recent advancements in measuring style similarity within diffusion models, the CSD~\cite{somepalli2024measuring} model introduces a methodology for extracting style descriptors. These descriptors can attribute the stylistic elements of a generated image to the specific images within the training dataset of a text-to-image model. Similar to CLIP~\cite{clip}, CSD is capable of producing distinctive style features, but with an emphasis on style discrimination. At each denoising step, we compute $x_0$ and its difference from the style reference at the CSD feature level. Subsequently, the derived gradient serves as style guidance to update the predicted noise.
\section{Experiments}

We have integrated our method with Stable Diffusion XL (SDXL)~\cite{podell2023sdxl}, adhering to the official implementation of InstantStyle~\cite{wang2024instantstyle}. For the image inversion process, we utilize the code provided by ReNoise~\cite{garibi2024renoise}. Our tile ControlNet~\cite{zhang2023adding} is trained from scratch. The original IP-Adapter~\cite{ye2023ip} is employed as the global image adapter for semantic preserving, and for style guidance, we have adopted the recently introduced CSD~\cite{somepalli2024measuring} model as our style extractor.

\begin{figure}[!h]
  \centering
  \includegraphics[width=\textwidth]{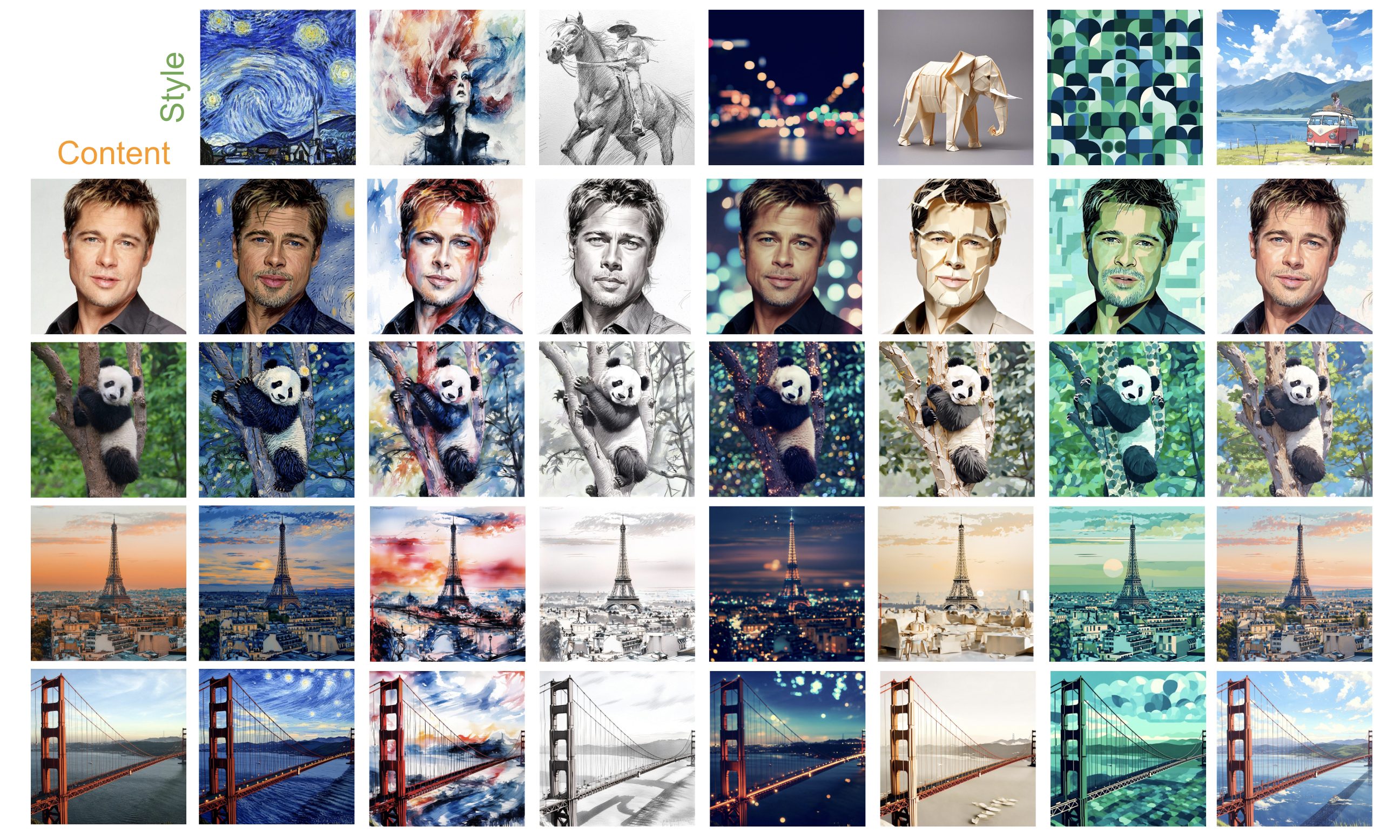}
  \caption{\textbf{Image-based image stylization results.} Given a content image and a style image, our training-free method can support content-preserving stylization. For human-centered stylization, we do not use any identity-preserving model for fair comparison.}
  \label{exp1}
\end{figure}

\subsection{Qualitative Results}


To ascertain the robustness and generalization capabilities of our method, we conducted an extensive series of style transfer experiments featuring a diverse range of styles applied to various content. Our focus was on preserving the content while optimizing the stylistic effect. Figure \ref{exp1} presents superior style transfer results across various subjects, demonstrating its robustness and versatility in adapting to diverse content. To ensure a fair comparison, we have not employed any identity-preserving models in our study.

\subsection{Comparison to Previous Methods}

\begin{figure}[!h]
  \centering
  \includegraphics[width=\textwidth]{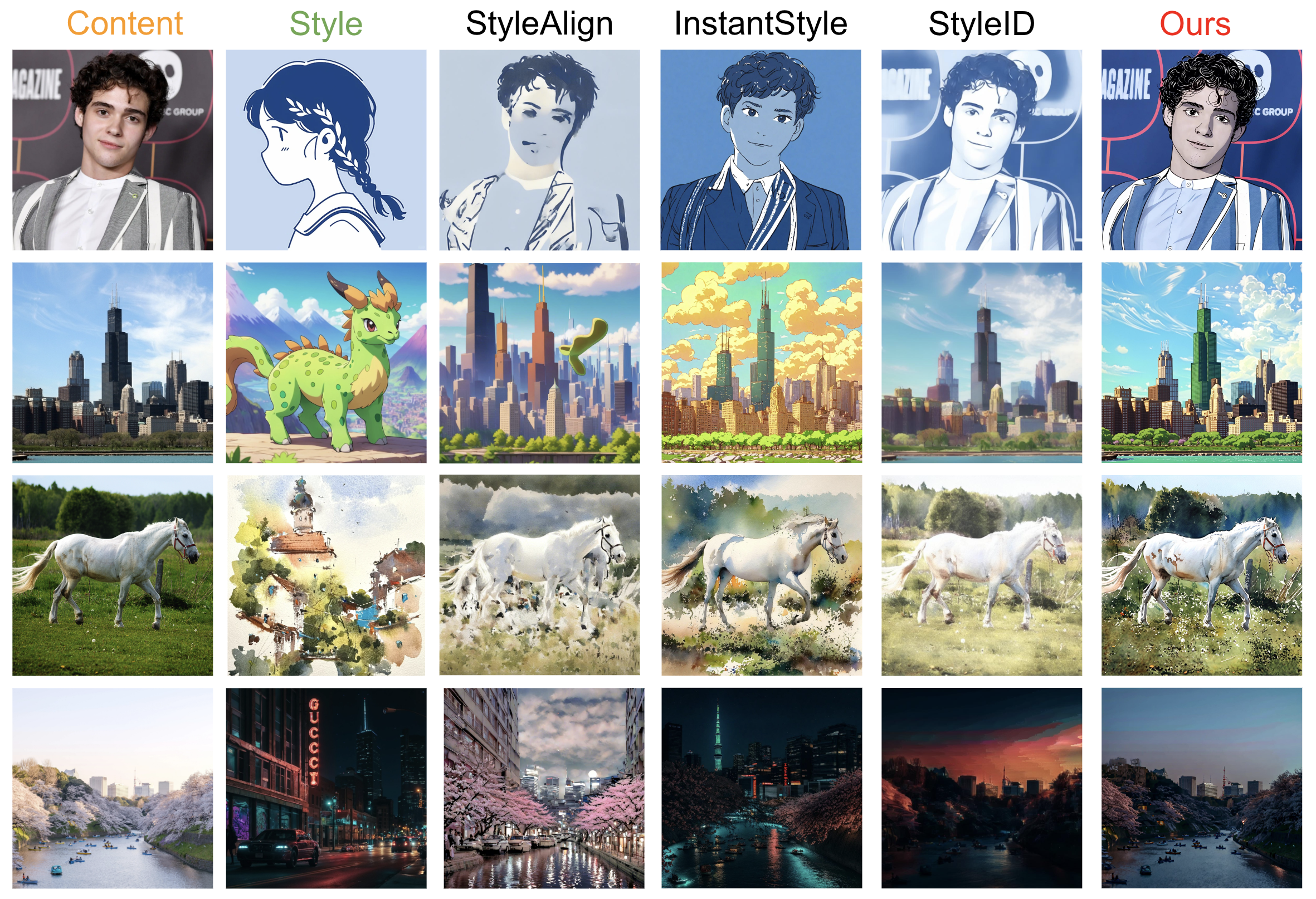}
  \caption{\textbf{Comparison to Previous Methods.} Except for StyleID which is already designed for content preserving, we utilize the official implementations of other works, integrating ControlNet for the purpose of spatial preservation.}
  \label{exp2}
\end{figure}

For establishing benchmarks, we compare our method against a suite of recent state-of-the-art stylization techniques, encompassing StyleAlign~\cite{hertz2023style}, InstantStyle~\cite{wang2024instantstyle}, and StyleID~\cite{chung2024style}. Among these, StyleID~\cite{chung2024style} distinguishes itself by emphasizing its capacity for content preservation. As shown in Figure \ref{exp2}, our approach achieves the best visual balance between enhancing stylistic effects and preserving the original content. In contract, prior works are either fail in preserving the integrity of the original content or to diminish the stylistic impact.

\subsection{Ablation Study}

In this section, we examine the impact of each incremental technique when applied to the style injection method, InstantStyle, as it pertains to our specific case. Specifically, we analyze the contributions of the initial content latent and tile ControlNet to the preservation of spatial composition, as well as the role of the global image adapter in maintaining semantic content. Lastly, we assess the impact of supplementary style guidance for stylistic compensation.

\begin{figure}[!h]
  \centering
  \includegraphics[width=\textwidth]{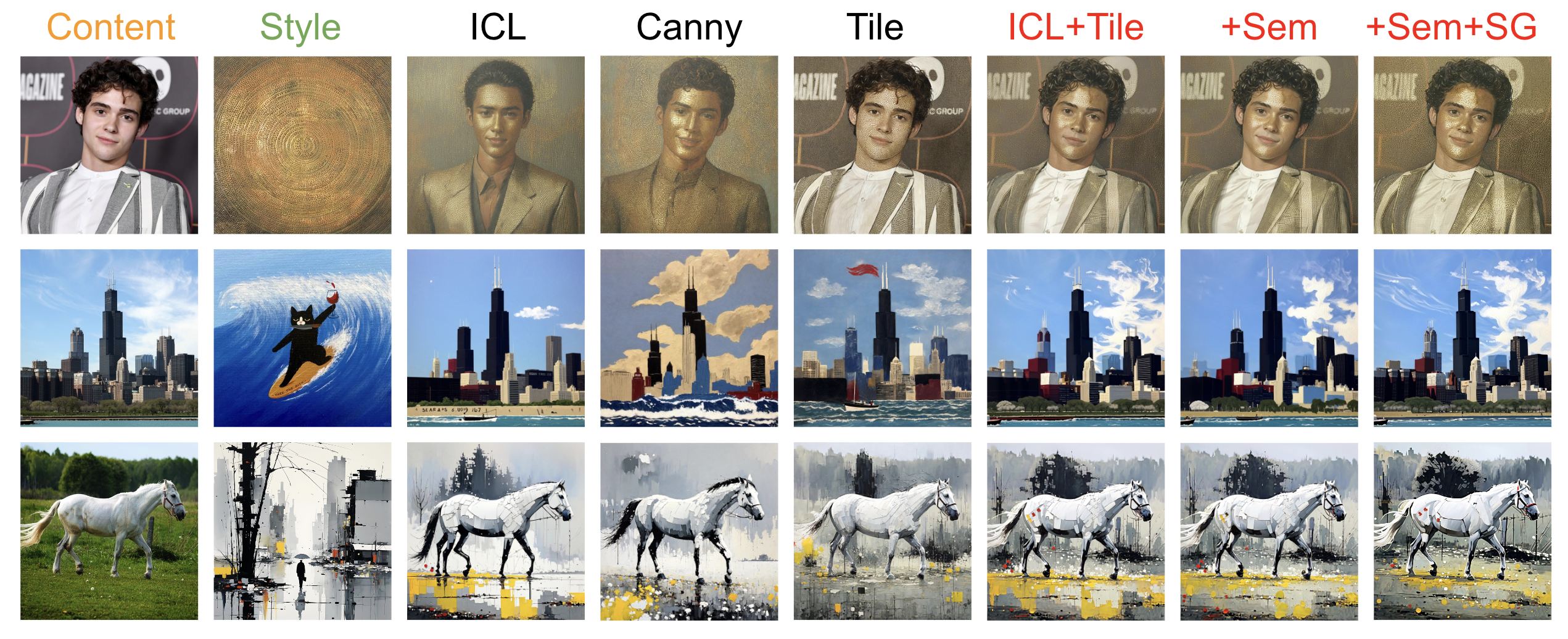}
  \caption{\textbf{Analysis of Sub-modules.} For the sake of brevity, we use the following abbreviations: Initial Content Latent (ICL), Canny ControlNet (Canny), Tile ControlNet (Tile), Semantic Preservation (Sem), and Style Guidance (SG).}
  \label{exp3}
\end{figure}

The results are shown in Figure \ref{exp3}. In terms of spatial structure preservation, our observations indicate that ControlNet is paramount in defining the overall structural framework, while the initial content latent plays a crucial role in retaining finer details. Regarding the selection of ControlNet, our comparative analysis reveals that Tile ControlNet significantly outperforms other conditions, such as Canny, for perserving content. For semantic preservation, we consider it a supplementary element to the text prompt. Its impact is minimal when a detailed description, \eg BLIP2~\cite{li2023blip} caption, has already been provided. Finally, our findings indicate that supplementary style guidance is instrumental in further enhancing the stylistic effect of the generated images, for example, in the first row of Figure \ref{exp3}, the stylistic details surrounding the mouth area are notably enhanced with the incorporation of style guidance.




\section{Conclusions and Future Work}

In this study, we leverage established methodologies rather than devising a new framework for content preservation in style transfer tasks. Our approach is anchored in the InstantStyle, a recent state-of-the-art technique for stylization. To augment its ability to preserve content fidelity, we first invert the content image to latent noise as initialization, which is designed to capture and retain the subtle details that are often overlooked in post-processes. Furthermore, We advocate the use of Tile ControlNet over other types, which facilitates in-place stylization without compromising the integrity of the original content. The two straightforward yet pivotal techniques are paramount for preserving the spatial structure of the content. We have also integrated a global image adapter to safeguard the semantic integrity of the content, which is particularly beneficial to avoid semantic drift in scenarios where the textual prompt is either absent or insufficiently descriptive. Lastly, we investigate the utility of the CSD model, which provides additional stylistic cues during the denoising phase, ensuring a more refined stylistic effect.

\subsubsection{Limitations.} As a pre-experimental project, our focus does not delve deeply into the interplay between content and style, instead, we only assess the practical utility of existing techniques in their applications. Several challenges still remain to be addressed. Firstly, the inversion process proves to be time-consuming, which could be a significant consideration for larger-scale applications. Secondly, we posit that the potential of Tile ControlNet has yet to be fully realized, suggesting that there is ample room for further exploration of its capabilities. Thirdly, the application of style guidance, while effective, demands substantial VRAM due to the accumulation of gradients across pixel space. This indicates a need for a more sophisticated approach to harnessing style signals efficiently.

\subsubsection{Future Work.} We are working on developing a more elegant framework to inject style without compromising the content integrity during the training phase, based on some observations in this report.

\clearpage
%
%
\bibliographystyle{splncs04}
\bibliography{egbib}



\end{document}